\documentclass[pdflatex,sn-mathphys-num]{sn-jnl}

\usepackage{graphicx}%
\usepackage{multirow}%
\usepackage{amsmath,amssymb,amsfonts}%
\usepackage{amsthm}%
\usepackage{mathrsfs}%
\usepackage[title]{appendix}%
\usepackage{xcolor}%
\usepackage{textcomp}%
\usepackage{manyfoot}%
\usepackage{booktabs}%
\usepackage{url}
\usepackage{subcaption}
\usepackage{algorithm}%
\usepackage{algorithmicx}%
\usepackage{algpseudocode}%
\usepackage{listings}%

\theoremstyle{thmstyleone}%
\theoremstyle{thmstyletwo}%

\theoremstyle{thmstylethree}%

\begin{document}

\title{BronchoLumen: Analysis of recent YOLO-based architectures for real-time bronchial orifice detection in video bronchoscopy}

\author[1]{\fnm{Yongchao} \sur{Li}}

\author*[1]{\fnm{Marian} \sur{Himstedt}}


\affil*[1]{\orgdiv{Faculty of Electrical Engineering and Computer Science}, \orgname{Technical University of Applied Sciences L\"ubeck}, \orgaddress{\street{M\"onkhofer Weg 239}, \city{L\"ubeck}, \postcode{23562}, \country{Germany}}}




\abstract{\textbf{Purpose:} Bronchoscopy is routinely conducted in pulmonary clinics and intensive care units, but navigating the complex branching of the respiratory tract remains challenging. This paper introduces BronchoLumen, a real-time YOLO-based system for detecting bronchial orifices in video bronchoscopy, aiming to assist navigation and CAD systems. The paper investigates if bronchial orifices can be robustly detected across image domains using state-of-the-art object detection and a limited set of public image data.

\textbf{Methods:} The study includes the description and comparison of YOLOv8, a widely adopted architecture, and YOLOv12, a more recent architecture integrating attention-based modules to improve spatial reasoning. Both models are trained and tested solely on publicly available datasets comprising different image domains. A comparison of both models is conducted based on the common metrics mAP@0.5 and mAP@0.5:0.9 with the latter emphasizing localization accuracy.

\textbf{Results:} For YOLOv8 we obtained a mAP@0.5 of 0.91 on an in-domain and 0.68 on a cross-domain test set. YOLOv12 achieved 0.84 and 0.68 respectively with slightly better localization accuracy with mAP@0.5:0.9 of 0.48 and 0.26 compared to YOLOv8 with 0.45 and 0.25. Challenges like motion blur and low contrast occasionally entailed uncertainties but the system demonstrated overall robustness in most scenarios.

\textbf{Conclusion:} BronchoLumen is an open-weight, YOLO-based solution for bronchial orifice detection offering high accuracy and efficiency across multiple image domains. While the more recent YOLOv12 achieves better localization accuracy, we observed a slightly worse precision. The models have been made publicly available to foster further research in bronchoscopy navigation. }


\maketitle

\newpage

\section{Introduction}
\begin{figure}[ht!]
    \centering
    \begin{subfigure}{0.45\textwidth}
      \includegraphics[width=\linewidth]{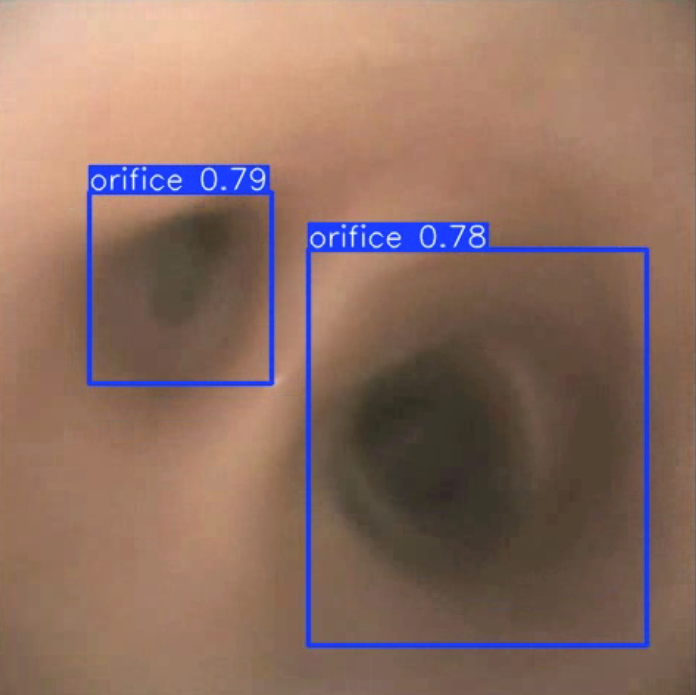}
      \caption{}
      \label{fig:first-page-figure-a}
    \end{subfigure}
    \begin{subfigure}{0.45\textwidth}
      \includegraphics[width=\linewidth]{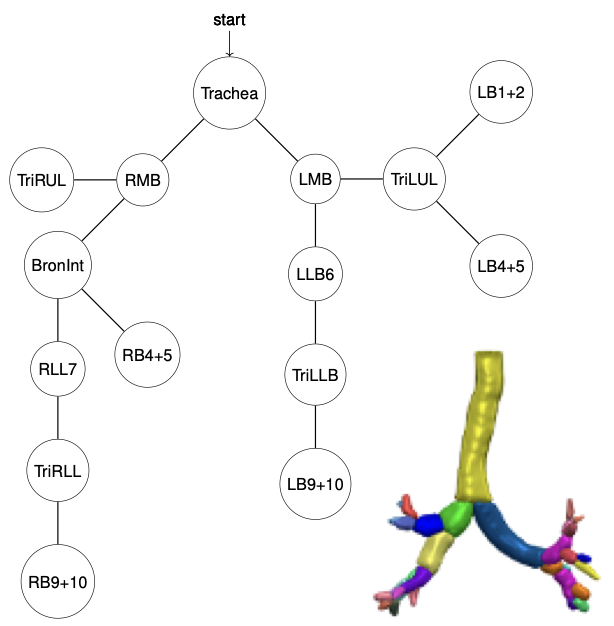}
      \caption{}
      \label{fig:first-page-figure-b}
    \end{subfigure}
    \caption{Bronchoscopy video frame augmented by bounding boxes with bronchial orifice detections (\ref{fig:first-page-figure-a}). Anatomical model of an airway tree comprising the first 4 branching generations (\ref{fig:first-page-figure-b}; image partly adapted from \cite{keuth_weakly_2022}).}
    \label{fig:first-page-figure}
\end{figure}

\footnotetext[1]{Code/Model available on: \url{https://github.com/mhimstedt/BronchoLumen}}
Video bronchoscopy is regularly conducted in pulmonary clinics as well as intensive care units (ICU), e.g. for lung biopsies and the treatment of acute respiratory problems. The human respiratory tract undergoes approximately 23 successive generations of branching (see also Fig. \ref{fig:first-page-figure-b}). Keeping track to the bronchoscope's position during interventions for recognizing anatomical landmarks is physically challenging for physicians \cite{yoo_deep_2021,zhao2024bronchocopilot,keuth2024airway}. State-of-the-art systems in pulmonary clinics utilize electromagnetic tracking and prior CT scans for navigation assistance \cite{eberhardt_lungpointnew_2010, nagao2004}. In recent years, the research community has pushed the state of the art in purely computer vision-based approaches using foundational work on bronchial orifice segmentation \cite{keuth_weakly_2022,bm-broncholc}, airway label prediction \cite{keuth2024airway, sganga_autonomous_2019}, depth image estimation \cite{xu2024depth, banach2021} and bronchoscopy image synthesis \cite{soliman2025bronchogan}. The detection of bronchial orifices as entrances to successive airway structures is of pivotal interest for several applications ranging from navigation assistance \cite{keuth2024airway,tian2024bronchotrack} and CAD systems \cite{bm-broncholc, yoo_deep_2021} to control of robotic bronchoscopy \cite{sganga_autonomous_2019,zhao2024bronchocopilot,zhang2024ai}. It enables topological navigation on a branch level for purely vision-based navigation - even in the absence of prior CT scans \cite{keuth2024airway, tian2024bronchotrack}. Computer vision models for objection detection are now widely accessible (e.g., \cite{redmon2016you}) and partly integrated in certified medical products, e.g. OLYSENSE (Olympus Corporation, Tokio, Japan) enabling diagnosis of lower and upper gastrointestinal (GI) diseases \cite{pacal2022efficient}. However, the detection of bronchial orifices has, to date, received relatively limited attention within the research community. Yiang et al. introduced BronchoTrack, a purely vision-based topological navigation system designed to detect and track bronchial orifices \cite{tian2024bronchotrack}. Despite showing promising results, the study lacks an independent evaluation of the integrated bronchial orifice detection. In addition, source code, models, and data were not made publicly available hampering the utilization in further applications. The limited availability of public video bronchoscopy datasets hampers the training of deep learning models. In this work, we include two public datasets \cite{bm-broncholc,sirglab-ds} for training a YOLO-based bronchial orifice detection model (see also Fig. \ref{fig:first-page-figure-a}).

\noindent \textbf{Contributions}:
\begin{itemize}
  \item Evaluation of orifice detection across different image domains on public datasets
  \item Open-source code and open-weight model to be shared with the research community
  \item Bounding box labels accompanying existing datasets fostering further research 
\end{itemize}

\section{Methods and data}\label{sec2}



\subsection{Bronchial orifice detection using YOLO}

We utilize the objection architecture YOLO \cite{redmon2016you} dividing an image into a grid and predicting bounding boxes and class probabilities directly in one network pass. This enables high inference speed with relatively low computational cost, making it well-suited for real-time applications. 
\noindent In this study, we included two object detection architectures: YOLOv8-M and YOLOv12-M using implementations of Ultralytics \cite{jocher2023yolov8}.

\subsubsection{YOLOv8-M} 
The YOLOv8-M architecture is composed of three main components: a Cross Stage Partial (CSP) backbone, a feature aggregation neck, and a decoupled detection head.
The backbone is built using Cross-Stage Partial Fusion (C2f) modules, which encourage feature reuse and efficient gradient flow. It replaces earlier CSP modules with a more compact implementation. Input images are first processed by a Focus block and downsampled via convolutional layers before being passed through a series of C2f blocks at multiple scales. The neck adopts a PANet-style structure for multi-scale feature fusion. Features from different stages are merged to preserve both spatial detail and semantic depth. The output consists of three scale-specific feature maps. YOLOv8 uses a decoupled head, where classification and regression branches are separated. It also operates in an anchor-free mode, directly predicting object centers and bounding box offsets, which simplifies training and reduces hyperparameters.

\subsubsection{Adapted YOLOv12-M}
\label{methods:yolov12}
The YOLOv12-M architecture builds upon the YOLOv8 backbone but introduces attention-based modules to enhance spatial reasoning and contextual representation.
Like YOLOv8-M, the backbone primarily uses C2f blocks for efficient feature extraction. In YOLOv12-M, A2 attention blocks are selectively inserted into deeper layers to model long-range dependencies and highlight salient anatomical features such as bronchial orifices. These modules are lightweight yet capable of enhancing global context awareness. The neck maintains a multi-scale fusion structure, similar to YOLOv8. In YOLOv12-M, attention-enhanced features are propagated through the neck, allowing for improved feature blending across scales. The detection head remains anchor-free and decoupled (separate classification/regression branches).

\noindent Note: To integrate YOLOv12-M into Ultralytics, we replaced its original R-ELAN structure with C2f modules. This adaptation preserved its attention-driven strengths while avoiding compatibility issues during training. Despite the addition of attention, the overall architecture remains lightweight and efficient enough for real-time inference.

\subsection{Datasets}

For training and testing our YOLO-architectures for bronchial orifice detection we utilize two public datasets: BM-BronchoLC \cite{bm-broncholc} (in-vivo) and SIRGLab-DS \cite{sirglab-ds} (in-vivo, ex-vivo, phantom). The SIRGLab-DS comprises video material of different domains, while BM-BronchoLC is limited to human in-vivo data. Tab. \ref{tab:dataset_stats} provides an overview of the utilized data. 

\begin{table}[ht!]
    \centering
    \caption{Dataset composition used for training and testing.}
    \label{tab:dataset_stats}
    \begin{tabular}{lcccc}
        \toprule
        \textbf{Subset} & \textbf{Dataset}               & \textbf{Domain} & \textbf{Resolution(s)} & \textbf{\# Images}\\ 
        \midrule
        Training        & \cite{bm-broncholc,sirglab-ds} & in-/ex-vivo     & 480/640             & 2900 \\
        Validation      & \cite{bm-broncholc,sirglab-ds} & in-/ex-vivo     & 480/640             & 438  \\
        Test Set 1      & \cite{bm-broncholc}            & in-/ex-vivo     & 640                 & 257  \\
        Test Set 2      & \cite{sirglab-ds}              & phantom         & 640                 & 362  \\
        \bottomrule
    \end{tabular}
\end{table}
\section{Results}

\subsection{Training} 

Both models were trained using the Ultralytics YOLO framework \cite{jocher2023yolov8} with the following hyperparameters:

\begin{table}[h]
\centering
\begin{tabular}{ll}
\hline
\textbf{Parameter} & \textbf{Value} \\
\hline
Epochs & 50 \\
Batch size & 16 \\
Image size & $640 \times 640$ \\
Optimizer & AdamW (default) \\
Initial learning rate & 0.001 (with cosine decay) \\
Learning rate schedule & Cosine decay \\
Warmup epochs & 3 \\
Weight decay & 0.01 \\
Loss function & Combined box, objectness, and classification loss \\
Confidence threshold & 0.25 (inference) \\
IoU threshold & 0.45 (non-maximum suppression and mAP) \\
\hline
\end{tabular}
\caption{Hyperparameters used for training}
\end{table}

\noindent All training runs used the same seed to ensure reproducibility. We used the same settings for both YOLOv8-M and YOLOv12-M to ensure a fair comparison.

\subsection{Quantitative results} 
\begin{figure}[ht!]
    \centering
    
    \begin{subfigure}{0.45\textwidth}
      \includegraphics[width=\linewidth]{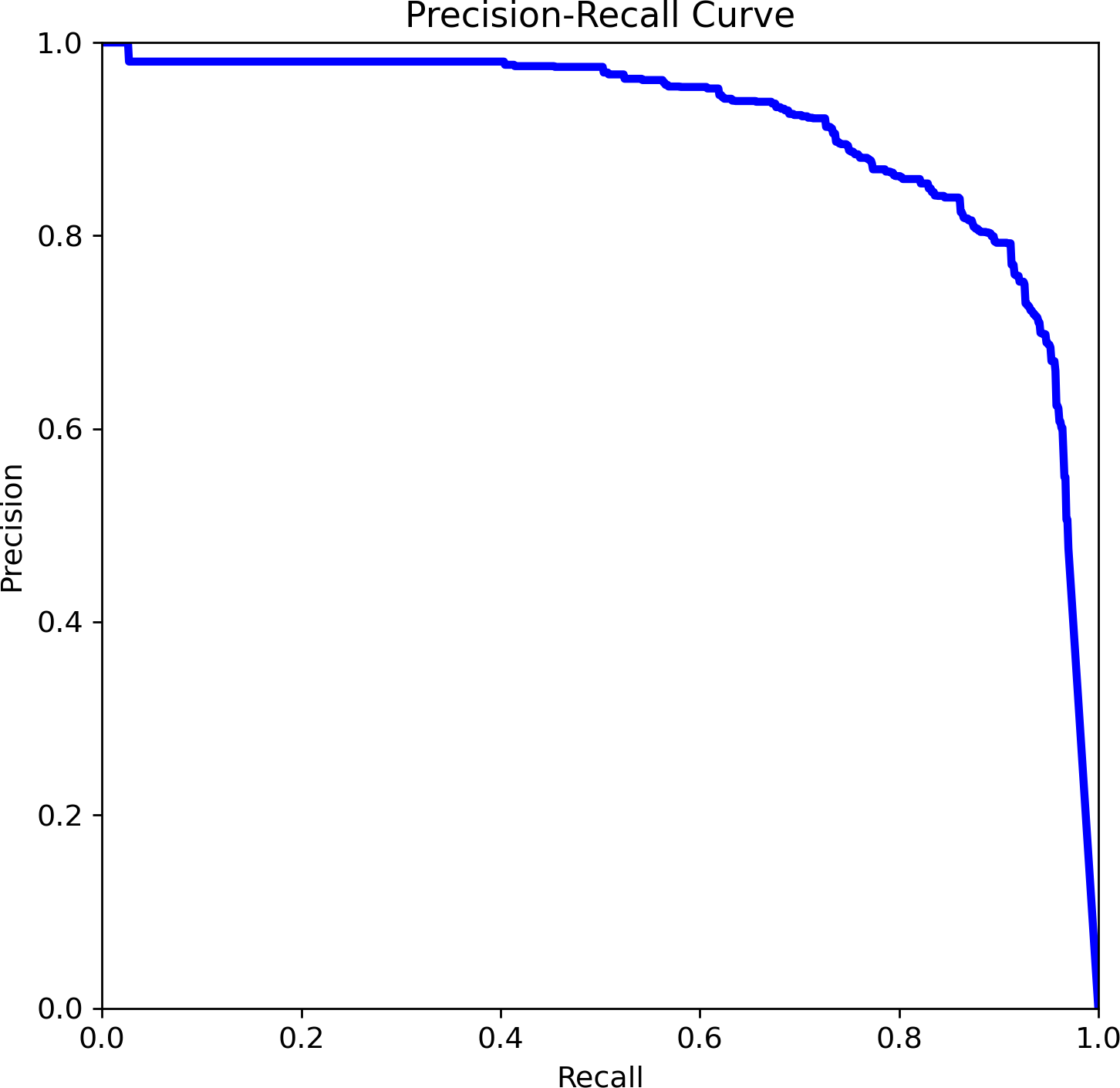}
      \caption{PR curve (Test Set 1)}
    \end{subfigure}
    \hfill
    \begin{subfigure}{0.45\textwidth}
      \includegraphics[width=\linewidth]{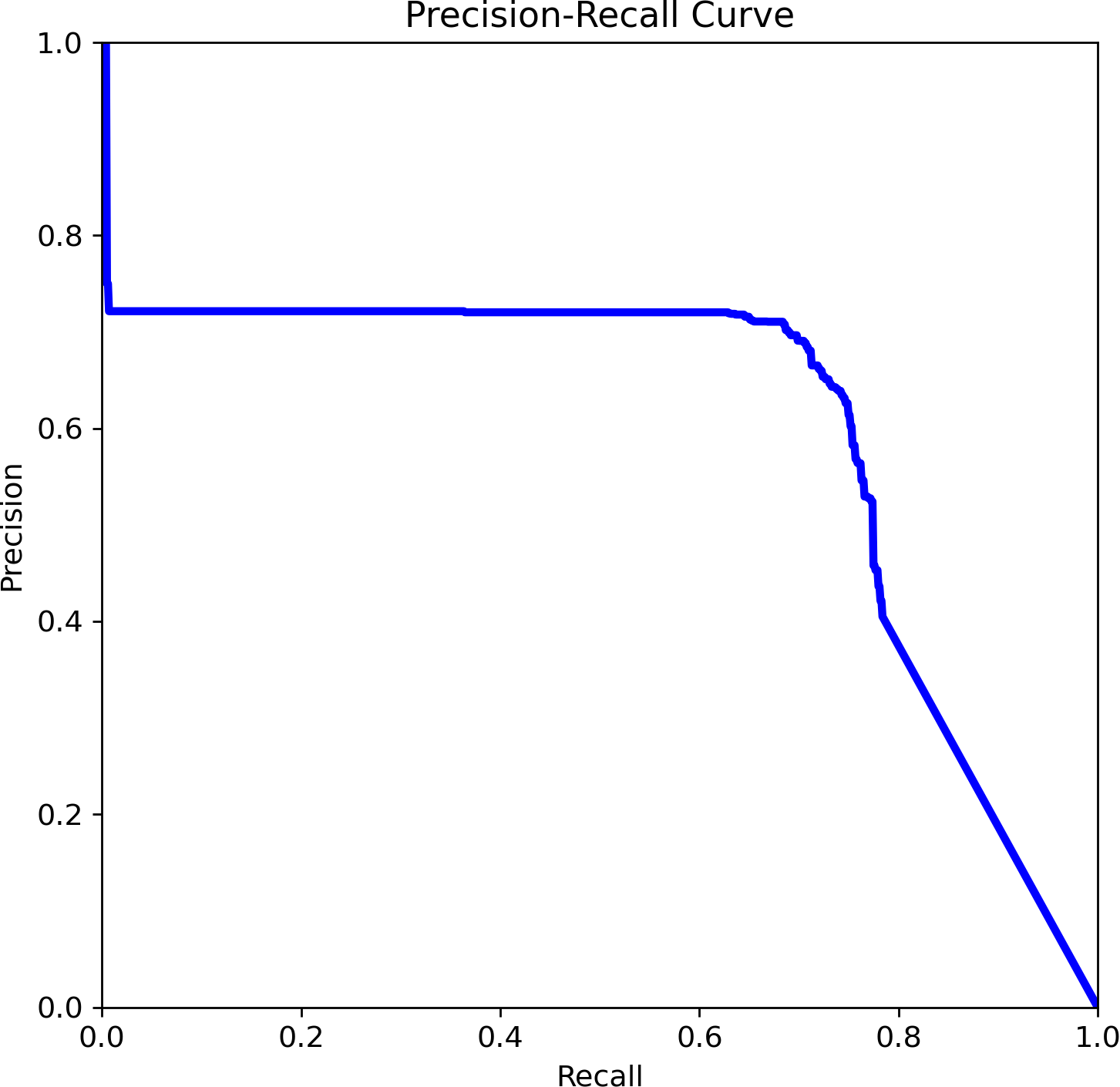}
      \caption{PR curve (Test Set 2)}
    \end{subfigure}
    \caption{Precision-recall (PR) curves obtained using YOLOv8 with the dataset split described in Tab. \ref{tab:dataset_stats}.}
    \label{fig:pr-curves}
\end{figure}

We trained and tested both models, YOLOv8-M and YOLOv12-M, on data split mentioned in Tab. \ref{tab:dataset_stats}. For the evaluation of our models we use common metrics: mAP@0.5 and mAP@0.5:0.95. The results obtained are shown in Tab. \ref{tab:yolo-comparison}. Precision recall curves for YOLOv8-M are shown in Fig. \ref{fig:pr-curves}. 

\begin{table}[ht!]
    \centering
    \caption{Comparison of YOLOv8-M and YOLOv12-M on test sets}
    \begin{tabular}{lcccc}
        \toprule
        Model & Dataset & Precision & mAP@0.5 & mAP@0.5:0.95 \\
        \midrule
        YOLOv8-M & Test Set 1 & 0.910 & 0.909 & 0.451 \\
        YOLOv8-M & Test Set 2 & 0.680 & 0.600 & 0.248 \\
        YOLOv12-M & Test Set 1 & 0.836 & 0.897 & 0.476 \\
        YOLOv12-M & Test Set 2 & 0.684 & 0.599 & 0.256 \\
        \bottomrule
    \end{tabular}
    \label{tab:yolo-comparison}
\end{table}

\subsection{Qualitative results} 
\begin{figure}[ht!]
    \centering
    \begin{subfigure}[t]{0.3\textwidth}
        \includegraphics[width=\linewidth]{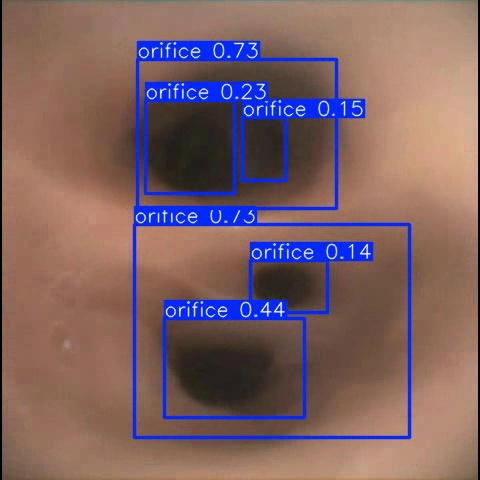}
        \caption{Nested structures}
    \end{subfigure}
    \begin{subfigure}[t]{0.3\textwidth}
        \includegraphics[width=\linewidth]{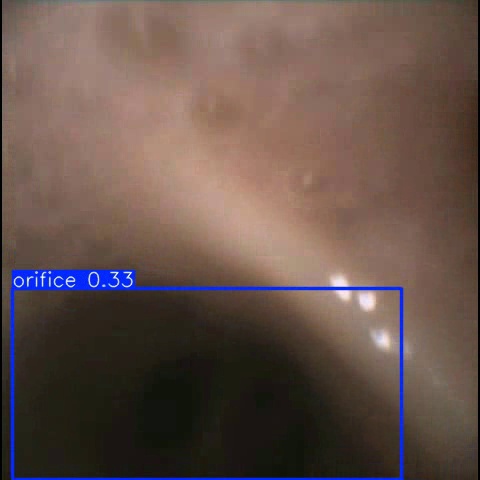}
        \caption{Motion blur}
    \end{subfigure}
    \begin{subfigure}[t]{0.3\textwidth}
        \includegraphics[width=\linewidth]{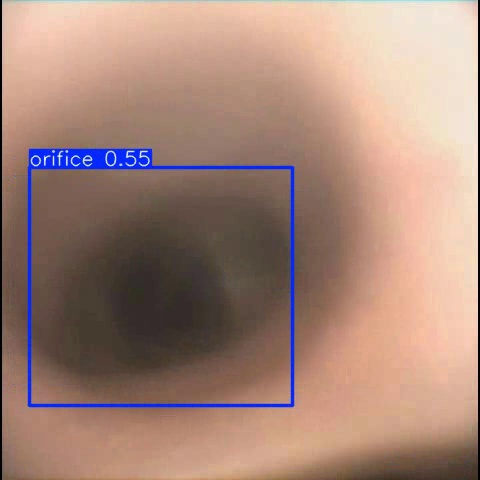}
        \caption{Low contrast}
    \end{subfigure}
    \caption{\textbf{Challenges}. Selected frames from an in-vivo bronchoscopy illustrating stable detection results throughout the sequence. Video material is originated from \cite{sirglab-ds}.}
    \label{fig:case_frames}
\end{figure}

Our test datasets comprise a number of challenging scenarios for lumen detection such as low contrast, motion blur and partial occlusion/visibility (see Fig. \ref{fig:case_frames}). The simultaneous detection of parental and child branches (nesting branches) is not erroneous but explicitly aimed and trained since this is beneficial for navigation tasks (see  Fig. \ref{fig:case_frames}a). 

\begin{figure}[ht!]
    \centering
    
    \begin{subfigure}[t]{0.34\textwidth}
      \includegraphics[width=\linewidth]{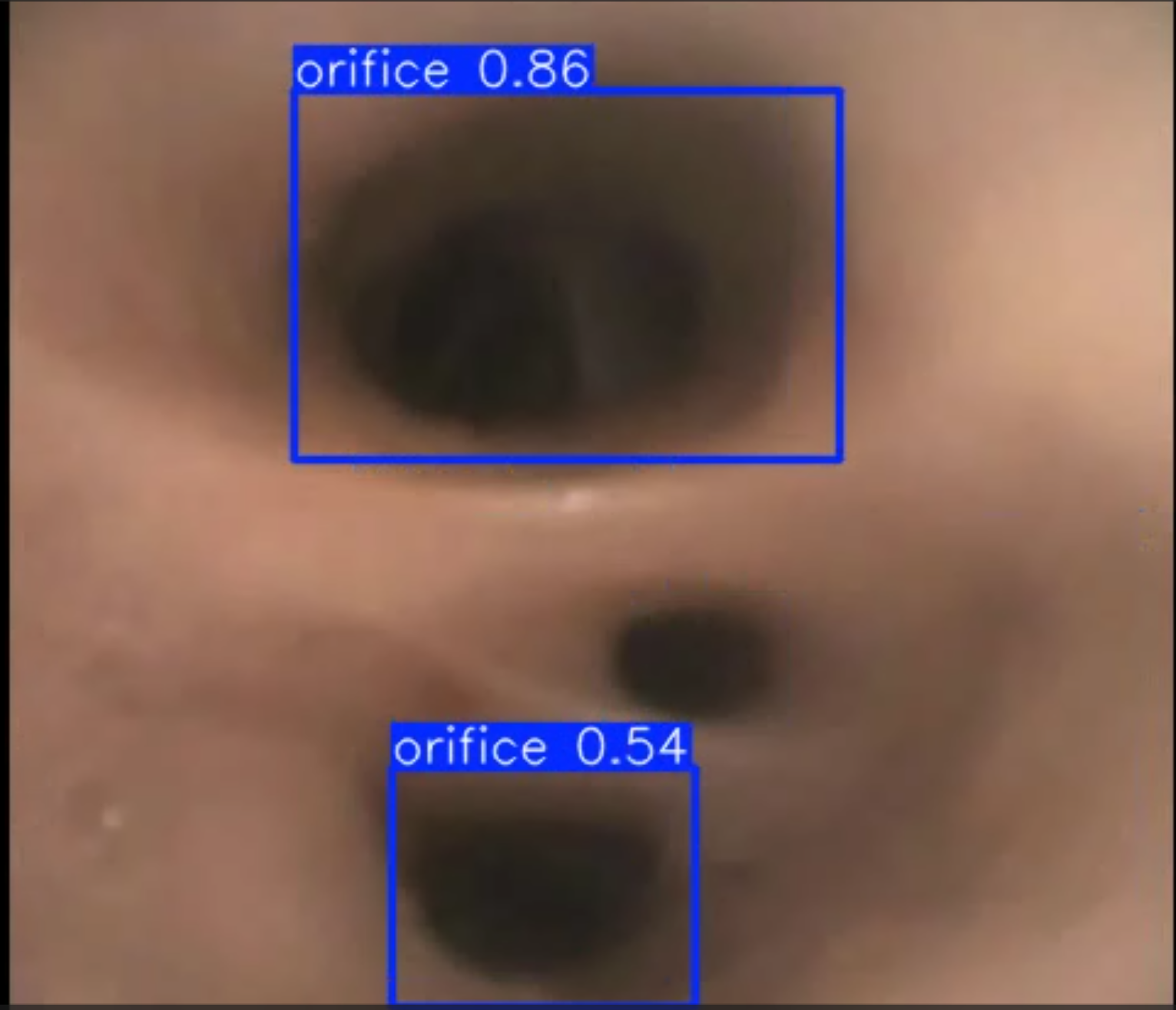}
      \caption{Missing orifice}
    \end{subfigure}
    \hfill
    \begin{subfigure}[t]{0.34\textwidth}
      \includegraphics[width=\linewidth]{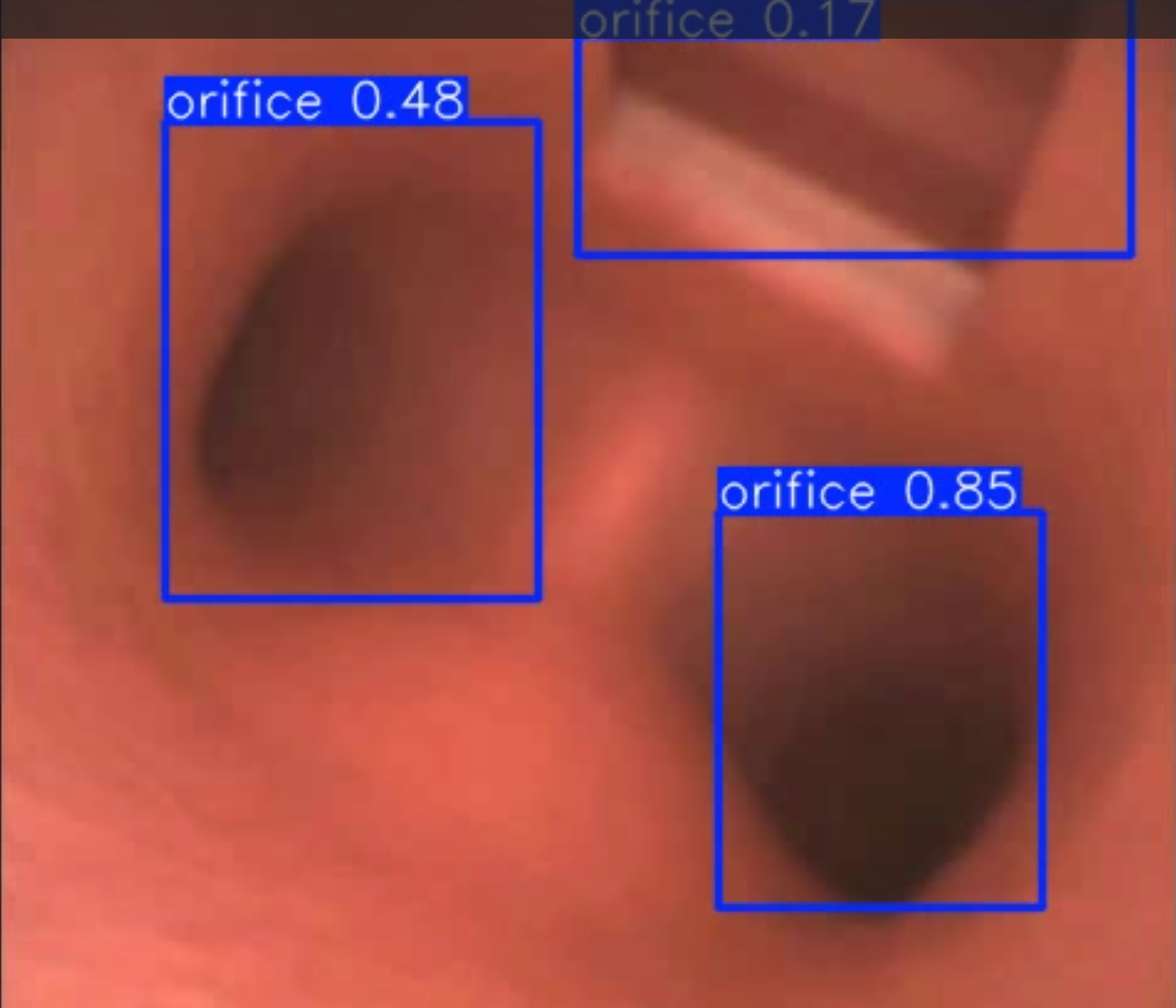}
      \caption{Erroneous detection}
    \end{subfigure}
    \begin{subfigure}[t]{0.29\textwidth}
      \includegraphics[width=\linewidth]{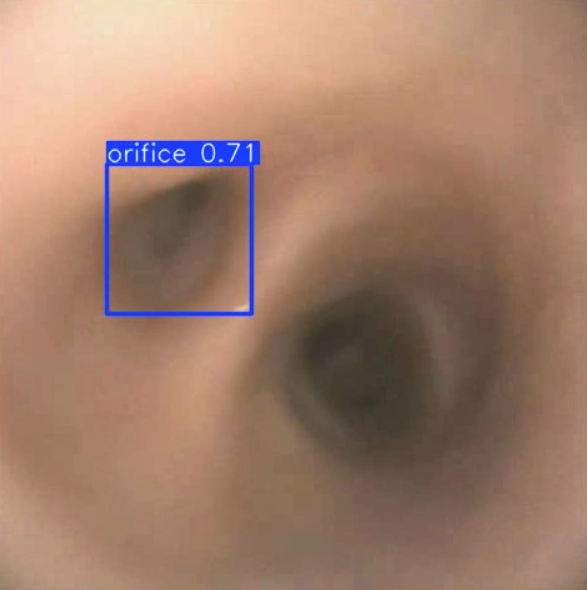}
      \caption{Missing orifice}
    \end{subfigure}
    \caption{\textbf{Failure cases}. Selected frames from mixed bronchoscopy domains showing false positive and false negative cases. Images obtained from \cite{sirglab-ds}.}
    \label{fig:negative-cases}
\end{figure}


\subsection{Inference time}
In addition to measuring detection accuracy, we evaluated the runtime performance of the \textbf{YOLOv8-M model} to determine its practicality for real-time clinical use. We performed video-based inference using an OpenCV script that processes each frame sequentially and records the total number of frames and overall processing time. On an \textbf{NVIDIA RTX 4060 GPU}, using a batch size of 1 and an \textbf{input resolution of $640 \times 640$}, the model achieved an average video \textbf{inference speed of 30.54 frames per second (FPS)}.

\subsection{Discussion}
Our quantitative analysis revealed that YOLOv12 slightly outperforms YOLOv8 in the localization accuracy (mAP@0.5:0.95) which confirmed the expectations due to the additional attention layers (see Sec. \ref{methods:yolov12}). However, a minor degradation of the overall precision (mAP@0.5) compared to YOLOv8 can be observed, too. When integrating BronchoLumen into applications, a trade-off must be considered: while improved localization accuracy may enhance stable path following in robotic bronchoscopy, higher precision (mAP@0.5) is generally preferable, such as in assistance systems with human operator-guided bronchoscopes.

Our model demonstrated high performance in many scenarios and is able to cope with numerous challenging situations (see Fig. \ref{fig:case_frames}). However, the limited availability of training data still entails false positive as well as false negative detections (see Fig. \ref{fig:negative-cases}). Motion blur, low contrast, and overexposed regions can still entail missing orifice detections. These issues were particularly noticeable in videos recorded under different lighting conditions or with less stable endoscope handling. Additionally, when the input resolution was reduced, a modest drop in visual clarity was observed, occasionally causing small orifices to go undetected despite improved inference speed. To further examine the impact of resolution, we conducted an experiment using a heavily downsampled version of the test set with an input size of $144 \times 144$. In this setting, the model's mAP@0.5 dropped sharply to 0.28. The detection results were often imprecise, and small or partially visible orifices were missed. This confirms that while lower resolutions can help with speed, reducing the input size below a certain threshold significantly impairs detection reliability. Fortunately, we rarely observed bronchial orifices being completely missed in a short video sequence. By incorporating tracking of orifices being detected once, we expect a more stable system which has also been investigated by \cite{tian2024bronchotrack}. 

\section{Conclusion}

We proposed an efficient bronchial orifice detection system based on the popular YOLO\cite{redmon2016you} architecture. In particular, we compared two variants: YOLO-v8 and YOLO-v12. While the former is a long-term stable and widespread architecture, YOLO-v12 incorporates novel advancements (attention layers) motivated by recent transformer architectures. YOLO-v12 improves localization precision but does not outperform YOLO-v8 in terms of detection accuracy. The inference time measured in our experiments confirms the real-time capability of our system leveraging orifice detection to be incorporated in future CAI and CAD systems. Our models were solely trained and tested on publicly available datasets simplifying the repeatability of the reported results. Source code for training and inference as well as model weights will be shared with the research community fostering further research and development for CAI/CAD systems in video bronchoscopy and to provide a fundament for further applications such as vision-based navigation in the human airways \cite{keuth2024airway,tian2024bronchotrack,sganga_autonomous_2019}. Models and software will be updated and shared with new datasets being publicly available. Both build up on the ultralytics ecosystem simplifying deployments over longer time.

\bibliography{sn-bibliography}

\end{document}